\pdfoutput=1

\documentclass[11pt]{article}

\usepackage[preprint]{coling}

\usepackage{times}
\usepackage{latexsym}

\usepackage[T1]{fontenc}

\usepackage[utf8]{inputenc}

\usepackage{microtype}

\usepackage{inconsolata}

\usepackage{graphicx}
\usepackage{amsmath}
\usepackage{algorithm}
\usepackage{algpseudocode}
\usepackage{authblk}

\usepackage{fancyhdr}

\usepackage[symbol]{footmisc}

\fancyhfoffset[L]{1cm} 
\fancyhfoffset[R]{1cm} 
\title{Beyond Visual Understanding\\Introducing PARROT-360V for Vision Language Model Benchmarking}

\author{
  \textbf{Harsha Vardhan Khurdula} \quad \quad
  \textbf{Basem Rizk} \quad \quad
  \textbf{Indus Khaitan} \\
  \textbf{Janit Anjaria}\footnote[2]{Advisor} \quad \quad \quad
  \textbf{Aviral Srivastava} \quad \quad
 \textbf{Rajvardhan Khaitan}
}

\affil{Redblock AI}

\begin{document}
\maketitle
\begin{abstract}
Current benchmarks for evaluating \textbf{Vision Language Models (VLMs)} often fall short in thoroughly assessing these models' abilities to understand and process complex visual and textual content. They typically focus on simple tasks that do not require deep reasoning or the integration of multiple data modalities to solve an original problem. To address this gap, we introduce the \textbf{PARROT-360V Benchmark}, a novel and comprehensive benchmark featuring $2487$ challenging visual puzzles designed to test VLMs on complex visual reasoning tasks. We evaluated leading models—\textbf{GPT-4o}, \textbf{Claude-3.5-Sonnet}, and \textbf{Gemini-1.5-Pro}—using PARROT-360V to assess their capabilities in combining visual clues with language skills to solve tasks in a manner akin to human problem-solving. Our findings reveal a notable performance gap: state-of-the-art models scored between \textbf{28 to 56\%} on our benchmark, significantly lower than their performance on popular benchmarks. This underscores the limitations of current VLMs in handling complex, multi-step reasoning tasks and highlights the need for more robust evaluation frameworks to advance the field.
\end{abstract}

\section{Introduction}
Vision Language Models (VLMs) have shown remarkable capabilities in integrating visual and textual data, excelling in tasks like image captioning and object recognition \cite{wang2024exploringreasoningabilitiesmultimodal}. 

The aspiration to create artificial intelligence that can seamlessly integrate into daily life—solving problems, performing tasks, and providing expert knowledge—has long been a driving force in technological advancement \cite{tandf}. Recent developments in VLMs have brought us closer to this vision, showcasing impressive abilities in understanding and generating both textual and visual data \cite{yang2024modelmergingllmsmllms}. Models like GPT-4o, Claude-3.5-Sonnet, and Gemini-1.5-Pro have set new standards in the field, while showcasing high performance for vision related benchmarks.

The rapid evolution of these models has sparked concerns. There is a growing fear that AI could replace human labor \cite{doi:10.1126/science.adj0998}. These fears are often based on hypothetical scenarios rather than current capabilities. Despite these apprehensions, it's crucial to critically assess whether these models truly perform at the levels claimed, especially in complex tasks that mirror real-world challenges.

Our benchmark, \textbf{PARROT-360V}, contributes to the evaluation of leading VLMs by focusing on step-by-step visual reasoning tasks. We aim to identify gaps between reported capabilities and actual performance, offering insights into specific areas where these models may underperform.

\section{Why A New Benchmark?}
Many commonly used benchmarks such as \textbf{MMMU} by \citet{yue2024mmmumassivemultidisciplinemultimodal}, \textbf{ChartQA} by \citet{masry2022chartqabenchmarkquestionanswering}, and \textbf{AI2D} by \citet{kembhavi2016diagramworthdozenimages} have been designed to evaluate VLMs on tasks that are limited in scope, such as basic image-text alignment or single-step reasoning. These benchmarks are typically straight forward, and models can at times overfit to the datasets, resulting in misleadingly high performance scores \cite{samuel2024datacontaminationdetectionmodern}. We aim to adequately test models on puzzles that require the skills of image-text alignment, multi-step reasoning and sequential logic handling. In particular order, they correspond to sub-tasks that are critical for real-world decision-making \cite{tu2024odeopensetevaluationhallucinations} and the analysis steps required.

\subsection{Challenges In Reproducibility For VLMs}
Reproducibility is a significant challenge in evaluating vision based tasks, especially when dealing with VLMs \cite{yang2024decomposecompareconsistencymeasuring}. Unlike purely textual models, vision models rely on visual input, which can be subject to variability in data preprocessing, annotation, and context \cite{wu2024highlightingsafetyconcernsdeploying}. This makes it difficult to replicate the exact conditions under which a model achieves specific results for other benchmarks. 

A lack of standardization in how input images are processed or prompts are structured can lead to discrepancies in model outputs when evaluated across different platforms or test environments \cite{anagnostidis2024susceptiblellmsinfluenceprompts}. Moreover, existing benchmarks, which often focus on answering a question posed at text within an image or on top of the content depicted in it (a graph), not adequately capturing the abilities of the model. The output for a question is heavily skewed by the size of data it was trained on. Rather benchmarking for VLMs should evaluate perception, and ability to use that information. And move away from the emphasis of answering multiple choice questions from an image, which could have easily been represented as a question in text \cite{yue2024mmmumassivemultidisciplinemultimodal}.

\subsection{A Fairer Evaluation Paradigm}
To ensure fairer comparisons, a benchmark should not just test how much knowledge a model has absorbed but should also evaluate how well it can perceive and follow instructions based on the visual inputs provided. This is where our PARROT-360V shines, as it requires models to integrate visual perception with textual reasoning, testing their ability to interpret, reason, and solve complex problems step by step, rather than regurgitating memorized knowledge \cite{NEURIPS2023_f26119b4}. 

This shift in focus is crucial for evaluating VLMs in a way that reflects their actual capabilities in real-world scenarios, where their performance must rely on accurate perception and decision-making rather than simply having been trained on vast quantities of data and determine its strengths towards being employed for automation involving visual tasks \cite{schwartz2023enhancingtrustllmbasedai}.


The \textbf{PARROT-360V} benchmark represents a step forward in addressing key challenges in reproducibility, data bias, and unfair comparisons in the field of VLMs. It provides a rigorous and fair benchmark for evaluating vision models based on their perceptual and reasoning abilities by employing Chain-of-Thought (CoT) \cite{wei2022chain} to plan how the model is going to solve the current puzzle, offering a clearer picture of how these models would perform in real-world applications \cite{wei2023chainofthoughtpromptingelicitsreasoning}.

\section{PARROT-360V Dataset}

\begin{figure}[t]
  \includegraphics[width=\columnwidth]{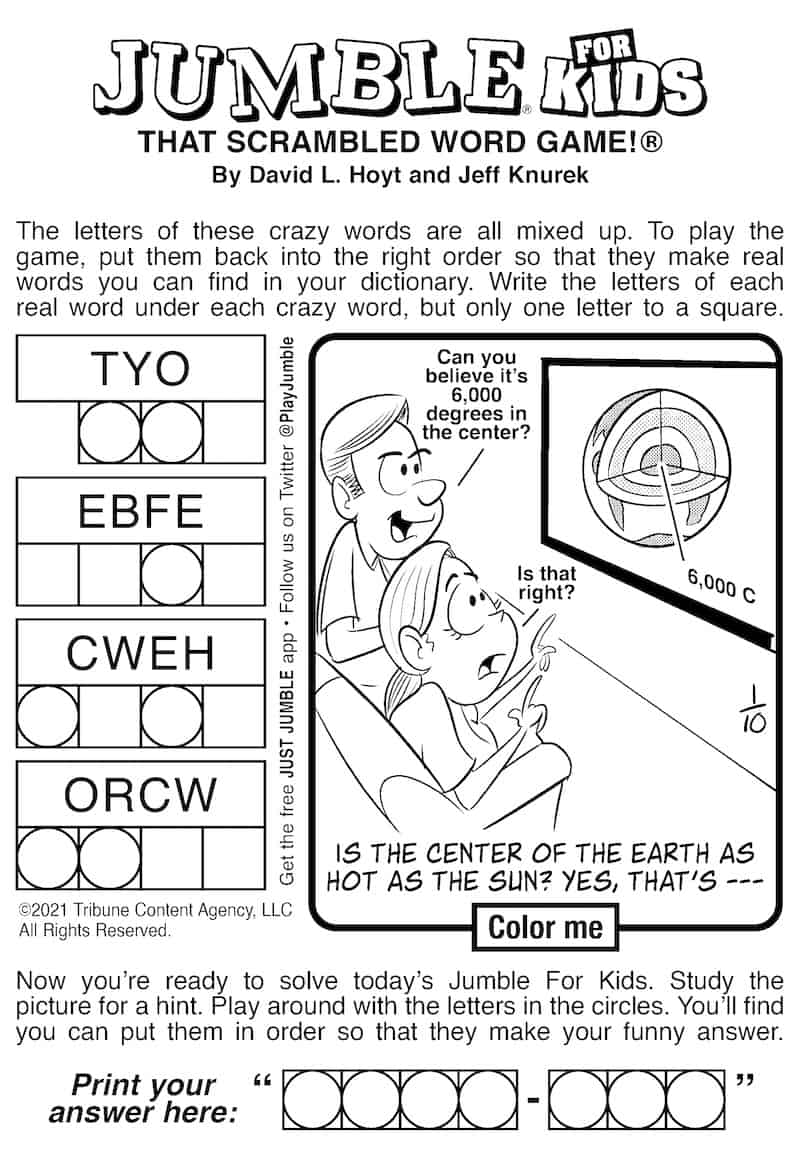}
  \caption{Sample from the PARROT-360V Dataset.}
  \label{fig:sampleFromDataset}
\end{figure}
The \textbf{PARROT-360V Benchmark} dataset was carefully curated by scraping Jumble puzzles from the internet to challenge VLMs in solving complex jumbles \cite{RedBlockAI_PARROT360V_2024}. Each scraped puzzle combines various elements representing an instance of gameplay, as shown in Figure \ref{fig:before_snap}, which serves as the input puzzle to the VLM. Furthermore, we extract additional features from the solved puzzle to obtain the ground truth labels, as shown in Figure \ref{fig:after_snap}.

The dataset corresponds to an established collection of puzzles, all of which are of the same format, with each containing a different set of clues/jumbled words, and a bonus clue with a visual component (associated with a cartoon) as shown in figure \ref{fig:sampleFromDataset}. The dataset is intended to evaluate not only language understanding but also visual perception and reasoning, making it a more rigorous test for these models.

\begin{table*}
  \centering
  \begin{tabular}{| lp{8cm} | lp{2cm} | }
    \hline
    \textbf{Field} & \textbf{Description} \\
    \hline
    Date & The date the puzzle was published. \\
    PDF File Path & The location of the puzzle in PDF format. \\
    Question Screenshot & A visual representation of the puzzle's clues. \\
    Answer Screenshot & A screenshot of the correct solution for comparison purposes. \\
    Clues (Clue\_1 to Clue\_4) & The four scrambled words presented as clues in the puzzle. \\
    Answers (Answer\_1 to Answer\_4) & The correct solutions to the jumbled clues. \\
    Visual Clue & A scrambled phrase or word that is solved using characters from the answers. \\
    Puzzle Answer & The correct solution to the bonus clue, derived from the characters circled in the answers. \\
    \hline
  \end{tabular}
  \caption{\label{tab:dataset_structure}
    Structure of the PARROT-360V Dataset, detailing the fields contained in each puzzle entry.}
\end{table*}

\subsection{Data Structure}
Each puzzle in the PARROT-360V dataset consists of the following features:
\begin{itemize}
    \item \textbf{Regular Clues}: These are scrambled words given as clues within the puzzle image, which the models must unscramble to find the correct words, and extract characters circled characters to form a bonus clue.
    \item \textbf{Visual Clue}: A cartoon or image that contains a visual hint. The models must interpret this image or other relevant information to form a bonus answer.
    \item \textbf{Answer Constraints}: The models are required to piece together specific letters (often circled in the image) from the unscrambled words to form the bonus answer.
\end{itemize}

The dataset contains $2487$ samples, with $14$ distinct features from the release date of the puzzle to the annotated answers as text within the dataset, as described in table \ref{tab:dataset_structure}.

\section{PARROT-360V Benchmarking Setup}
We applied our benchmark on three state-of-the-art VLMs—\textbf{GPT-4o}, \textbf{Claude-3.5-Sonnet}, and \textbf{Gemini-1.5-Pro}. Requiring the models to integrate both visual and textual information to arrive at correct answers, the evaluation environment is designed to simulate a realistic puzzle solving environment:

\begin{itemize}
    \item Input: Images of jumbled word puzzles, including visual clues (circled letters, and cartoon characters).
    \item Task: The models are required to solve the scrambled words, interpret the visual clues, and synthesize the final solution by forming bonus answers from the circled letters in the puzzle. While framing context from the posed question in relation to the cartoon (as shown in the algorithm \ref{prompt_algo}).
    \item Metrics: We measured the correctness of their responses and the proportion of hallucinations, where the model incorrectly used characters that were not part of the given clues. Additionally, we calculated each model's overall performance across multiple dimensions—accuracy, sequential performance (evaluation of intermediate steps involved), and hallucination rate.
\end{itemize}

\begin{algorithm*}[h]
\caption{TASK: Solve Jumble Puzzle}
\label{prompt_algo}
\begin{algorithmic}[*]
\State \textbf{Input:} Image containing jumbled words $W_1, W_2, W_3, W_4$, circled letter positions, bonus section, and a visual clue.
\State \textbf{Output:} Unscrambled words $U_1, U_2, U_3, U_4$, reasoning behind the bonus answer $B$
\State \textbf{Planning Phase:}
\State Recognize that each word needs to be unscrambled and circled letters extracted to form the bonus clue.
\State \textbf{Execution Phase:}
\For{$i = 1$ to $4$}
    \State Unscramble $W_i$ to get $U_i$
    \State Extract the circled letter from $U_i$ at position $P_i$
\EndFor
\State Concatenate circled letters to form bonus clue $C$
\State Unscramble $C$ to get the bonus answer $B$ while considering the posed question and visual clue.
\end{algorithmic}
\end{algorithm*}

To mitigate the issue of potential data contamination, we ensured that the setup used was entirely novel and tests the models for real world task, that cannot be reproduced by simply using its knowledge base. Rather we prompt the VLM to explicitly/step-by-step address the required tasks: identify characters, plan on how it is going to handle the bonus clue, and solve with the entire puzzle. 

One of the core challenges presented by the PARROT-360V benchmark is for VLMs to identify the circled letters within the image clues as shown in \ref{fig:sampleFromDataset} for a model. These letters are crucial for solving the bonus clue, and failure to correctly identify these letters often results in hallucinations or incorrect answers from the models.

In addition to recognizing circled characters, models are expected to interpret visual information from the cartoon or accompanying image (Figure \ref{fig:sampleFromDataset}). This requires not only extracting individual characters but understanding the broader context of the image to form a coherent solution to the puzzle. These puzzles that require visual understanding, multi-step reasoning, and sequential logic challenge the perception that current VLMs excel in complex, real-world tasks. This plays a critical role in understanding the limitations of these models in tasks that go beyond simple image-text QA \cite{yue2024mmmumassivemultidisciplinemultimodal}.

\begin{table*}
  \centering
  \begin{tabular}{lccccc}
    \hline
    \textbf{Model}           & \textbf{MMMU} & \textbf{Mathvista} & \textbf{AI2D} & \textbf{ChartQA} & \textbf{Average Performance} \\
    \hline
    GPT-4o              & 0.69 & 0.64 & 0.94 & 0.86 & \textbf{0.78} \\
    Claude 3.5 Sonnet   & 0.68 & 0.68 & 0.95 & 0.91 & \textbf{0.80} \\
    Gemini 1.5 Pro      & 0.62 & 0.64 & 0.81 & 0.81 & \textbf{0.72} \\
    \hline
  \end{tabular}
  \caption{\label{benchmark-table}
    Performance of different models across various benchmarks.
  }
\end{table*}

\subsection{Metrics}
To quantify the models' performance on PARROT360V, we developed a scoring system that assigns \textbf{weights} to each component of the puzzle. The scoring system is designed to reflect the importance of each task and to penalize omissions or incorrect answers appropriately.

Scoring components each puzzle consists of:
\begin{itemize}
    \item \textbf{Four Scrambled Words}: Each worth 10 points. (There are four scrambled words/clues, within each puzzle thus a candidate can score 40 points at the most in this section.)
    \item \textbf{Synthesizing Answers to Extract Key Characters}: Worth 10 points. (Each unscrambled clue serves as a distinct answer. Certain characters are circled within these answers; concatenating these circled characters provides the final bonus clue.)
    \item \textbf{Solution to the Puzzle}:  Worth 20 points. (Using the extracted characters and interpreting the cartoon, VLMs gather and synthesize information to solve the puzzle.)
\end{itemize}

The total possible points for each puzzle are 70 points.

\subsection{Scoring Methodology}

For each VLM's attempt at solving a puzzle, we applied the following evaluation criteria:
\begin{itemize}
    \item Correct Answer: If the model's answer matches the  labeled correct answer exactly (case-insensitive), it receives full points for that component\footnote{Clue or One of the sequential tasks involved in solving the given puzzle.}.
    \item No Answer or Incorrect Answer: If the model provides no answer or an incorrect answer, it receives a penalty of -5 points for that component.
    \item Negative Score Adjustment: If the total points earned for a puzzle are negative due to penalties, the score is clipped to zero.
    \item Normalization: The total points earned are divided by the total possible points (70) to obtain a performance score between 0 and 1, rounded to two decimal places.
\end{itemize}

Let:

\begin{itemize}
    \item $W_i$ be the weight for component $i$.
    \item $A_i$ be the model's answer for component $i$.
    \item $C_i$ be the correct answer for component $i$.
    \item $T$ be the total possible points (70).
\end{itemize}

For each component:

\[
P_i =
\begin{cases} 
      W_i & \text{if } A_i \text{ is exactly } C_i, \\
      -5 & \text{if } A_i \text{ is incorrect or missing}.
\end{cases}
\]

Total Points Earned:

\begin{equation}
P_{\text{total}} = \sum_i P_i
\end{equation}

Clipped Total Points Earned:

\begin{equation}
P_{\text{adjusted}} = \max(0, P_{\text{total}})
\end{equation}

Performance Score:

\begin{equation}
PARROT360V_{\text{Score}} = \frac{P_{\text{adjusted}}}{T}
\end{equation}

Hallucination within PARROT-360V benchmarking is relates to the frequency with which models introduced information not present in the input. And Hallucination rate is the error rate, i.e. the proportion of incorrect predictions given by an VLM:

\begin{equation}
Hallucination Rate = 1 - (PARROT360V_{\text{score}})
\end{equation}

\section{Results}

The evaluation of GPT-4o, Claude-3.5-Sonnet, and Gemini-1.5-Pro on our benchmark highlights the significant limitations of existing benchmarks in capturing true multimodal reasoning abilities. Unlike tasks found in common benchmarks like \textbf{MMMU}, \textbf{MathVista}, or \textbf{ChartQA} (as shown in the table \ref{benchmark-table}), PARROT-360V places special emphasis on complex, multi-step reasoning involving visual puzzles. This difference is reflected in the sharp decline in performance when these models are tested on our benchmarking dataset.

\begin{figure}[t]
  \includegraphics[width=\columnwidth]{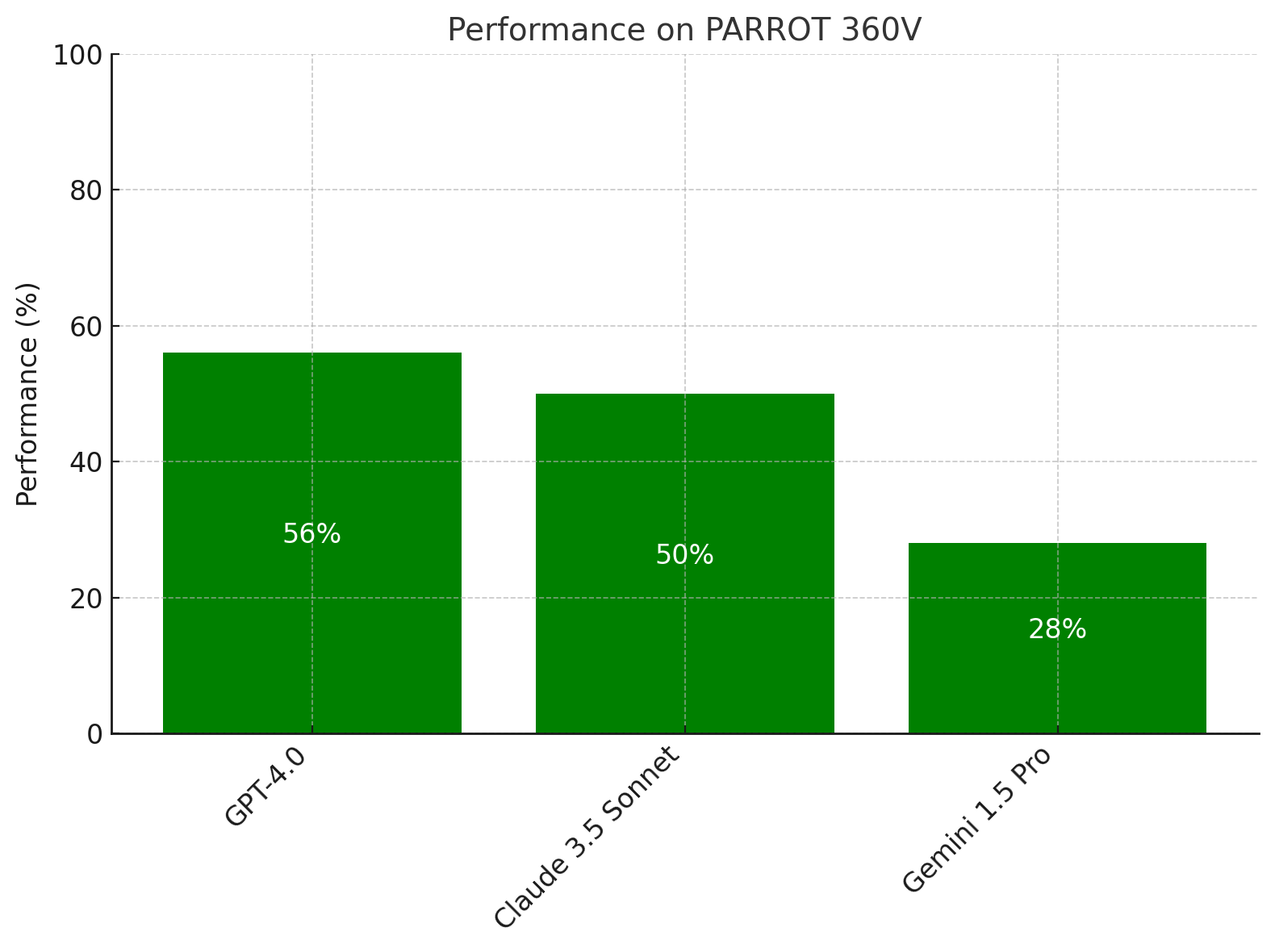}
  \caption{Performance of State-of-The-Art VLMs on PARROT-360V}
  \label{fig:360V}
\end{figure}

On benchmarks such as \textbf{MMMU} and \textbf{MathVista}, GPT-4o and Claude-3.5-Sonnet achieved high scores of 0.69 and 0.72, respectively (Table \ref{benchmark-table}), mainly because these tasks focus on simple image-text alignment or basic reasoning. In these tests, the image often serves merely as a backdrop to a question that could just as easily be presented as pure text, reducing the need for genuine visual understanding.

PARROT-360V, by contrast, involves complex tasks like word unscrambling, bonus clue extraction, and interpreting visual elements, all requiring deep integration of visual and textual information. GPT-4o’s accuracy dropped significantly to \textbf{0.57}, Claude-3.5-Sonnet's to \textbf{0.50}, and Gemini-1.5-Pro's to \textbf{0.28} (Figure \ref{fig:360V}) in this more challenging setup, demonstrating how existing benchmarks fail to reflect the complexity required for real-world tasks.

\subsection{Visual Perception Failures}
In tasks such as identifying circled letters with-in the puzzle which is an image (Figure \ref{fig:sampleFromDataset}), all three candidate models struggled. For instance, Gemini-1.5-Pro exhibited a high hallucination rate of \textbf{72\%}, largely due to its inability to accurately recognize and use visual inputs. This stands in contrast to simpler benchmarks like \textbf{AI2D}, where models face straightforward visual questions with limited need for complex image interpretation \cite{kembhavi2016diagramworthdozenimages}. Often the models also failed to synthesize visual scrambled text effectively. The challenge of extracting circled letters and forming a correct bonus clue required higher-order detail to reasoning, which is not typically tested in conventional benchmarks.

\subsection{Hallucination Issues}
All three VLMs exhibited frequent hallucinations during evaluation, particularly when trying to derive answers from visual cues. While tasks in \textbf{ChartQA} or \textbf{MathVista} are often solved by applying memorized data or pattern recognition, Our benchmark exposed the models' limitations in handling dynamic, real-time visual information. GPT-4o had a hallucination rate of \textbf{43\%}, Claude-3.5-Sonnet with \textbf{50\%}, and Gemini-1.5-Pro with \textbf{72\%} as shown in the figure \ref{fig:360V}, thus proving how heavily models depend on structured data rather than real reasoning from raw inputs.

\section{Discussion}
When we compared the results of our proposed benchmark to benchmarks like MMMU, ChartQA, MathVista, and AI2D, it became clear that these existing benchmarks don’t truly test a model’s ability to reason through complex, real-world visual problems. In benchmarks like MMMU, models often achieve high scores (e.g., GPT-4o scoring 0.69 as shown in table \ref{benchmark-table}) because the tasks typically involving static image-text alignment or basic pattern recognition. These benchmarks don’t require the models to think, but rather to retrieve memorized information or recognize patterns from training data. In essence, they reduce the challenge to answering questions that could just as easily be text-based.

In contrast, our framework challenges the models with multi-step reasoning and visual puzzles that demand a higher level of understanding. Models can’t just rely on large datasets or pre-learned patterns; they need to synthesize information from both text and images. For example, tasks like identifying circled letters in images and using them to solve a bonus clue are far more reflective of real-world complexity than simple image-caption matching. When we saw models like Gemini-1.5-Pro struggle with hallucinations (72\% rate as seen in figure \ref{fig:360V}), it was a clear indication that they’re not truly equipped for these kinds of tasks—yet these are the tasks that matter when it comes to applying AI in fields like healthcare or automation.

One of the biggest takeaways from PARROT-360V was that the models performed significantly worse on our benchmark—with performance scores as low as \textbf{28\%} for Gemini-1.5-Pro—compared to traditional benchmarks (Figure \ref{fig:360V}). In short, current benchmarks are giving us an incomplete and often inflated view of what these models can really do. The models' performance drop on PARROT-360V proves that while they might excel at answering questions from pre-learned data, they struggle when it comes to reasoning through complex, multi-step visual tasks. To move forward, we need benchmarks like our PARROT-360V that challenge these models to think and reason, not just recognize or recall.

\section{Conclusion}
With \textbf{PARROT-360V}, we aim to push the boundaries of how we evaluate VLMs by focusing on real-world tasks that demand visual perception, multi-step reasoning, and instruction-following. We saw that traditional benchmarks are not enough. They tend to focus on simpler tasks like image-text alignment and QA, which doesn’t truly challenge the models' ability to understand and process both visual and textual data together. In contrast, PARROT-360V makes models tackle tasks that require actual reasoning and visual integration, such as solving word puzzles with visual clues.

Our findings reveal that GPT-4o (56\%), Claude-3.5-Sonnet (50\%), and Gemini-1.5-Pro (28\%) struggle on our benchmark when handling complex, real-world tasks. This performance gap underscores the need for a more reliable benchmark, which PARROT-360V aims to provide.

\section{Limitations}
While \textbf{PARROT-360V} introduces a fresh approach to evaluating VLMs, we recognize that there are areas where further refinement can enhance its robustness. One aspect we’ve observed is the task complexity. Puzzles within the proposed benchmarking dataset are intentionally challenging, designed to test multi-step reasoning and visual perception. However, there are instances where the complexity may obscure whether a model’s failure is due to genuine reasoning difficulties or simply the task's intricacy. As we move forward, maintaining the right balance in task difficulty will help ensure we are accurately measuring a model’s reasoning capabilities.

Another focus is visual perception, as models must interpret visual clues and recognize circled letters. We aim to separate perception from reasoning to ensure fair evaluation. Lastly, to address data contamination, we will regularly update the benchmark with new tasks to test models on unseen data.


\bibliography{custom}

\appendix

\section{Scraping And Puzzle Curation}
\label{sec:appendix}

\begin{figure}[h]
  \includegraphics[width=\columnwidth]{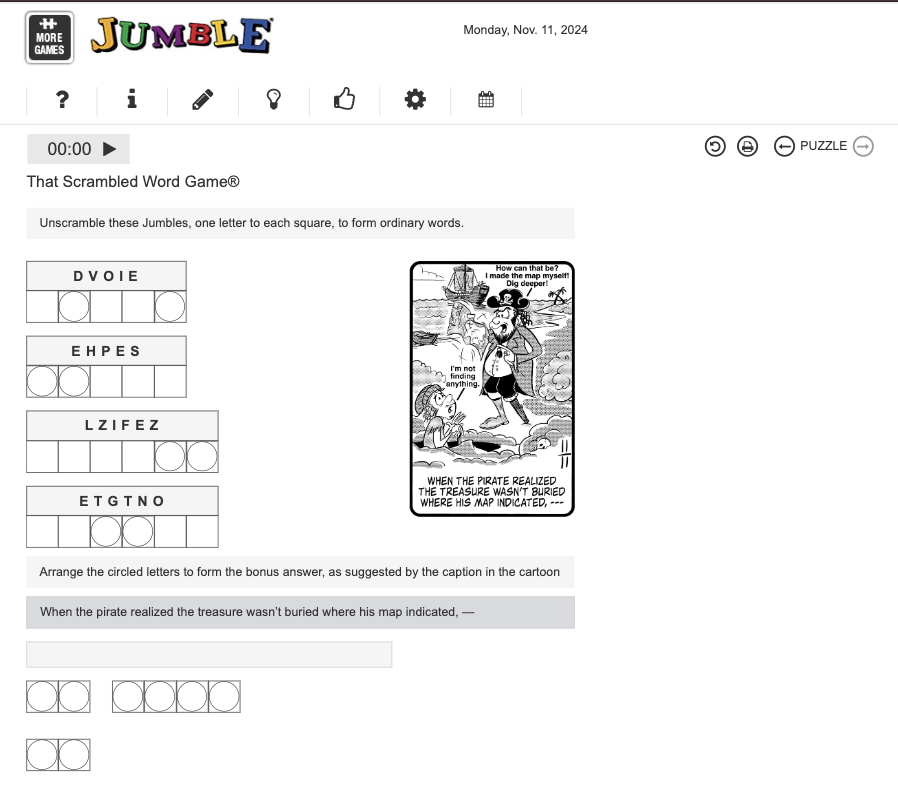}
  \caption{Snapshot of the Puzzle.}
  \label{fig:before_snap}
\end{figure}

\begin{figure}
  \includegraphics[width=\columnwidth]{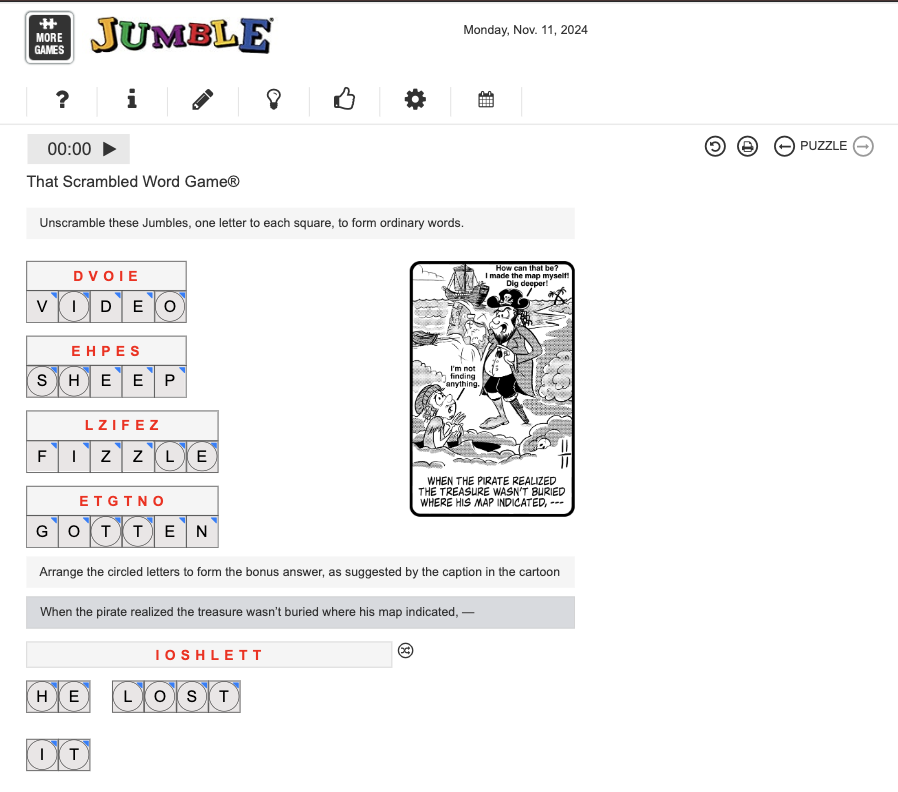}
  \caption{Snapshot of the Solved puzzle.}
  \label{fig:after_snap}
\end{figure}

\end{document}